\documentclass[letterpaper, 10 pt, conference]{ieeeconf}  

\usepackage{censor}                 
\usepackage{url,hyperref,lineno,microtype,subcaption}
\usepackage{cite}
\usepackage{amsmath,amssymb,amsfonts}
\usepackage[compatibility=false]{caption}
\usepackage{graphicx}
\usepackage{algorithm}
\usepackage{algpseudocode}
\algrenewcommand\alglinenumber[1]{\scriptsize #1:}
\usepackage{booktabs}
\usepackage{tabularx}
\usepackage{multirow}
\usepackage{array}

\usepackage{amsmath,amsfonts,amssymb}

\newcommand{\vs}{vs.\ }
\newcommand{\ie}{{\em i.e.\ }}

\newcommand{\et}{{\em et al.\ }}

\newcommand{\lip}{\mbox{$\langle$}}
\newcommand{\rip}{\mbox{$\rangle$}}

\newcommand{\lar}{\ensuremath{\;\,\leftarrow\;\,}} 

\newcommand{\tb}{\mbox{$\bf t$}}

\newcommand{\vb}{\mbox{$\bf v$}}
\newcommand{\wb}{\mbox{$\bf w$}}
\newcommand{\xb}{\mbox{$\bf x$}}

\newcommand{\AB}{\mbox{$\bf A$}}

\newcommand{\EB}{\mbox{$\bf E$}}

\newcommand{\IB}{\mbox{$\bf I$}}

\newcommand{\TB}{\mbox{$\bf T$}}

\newcommand{\AC}{\mbox{$\mathcal A$}}
\newcommand{\BC}{\mbox{$\mathcal B$}}

\newcommand{\GC}{\mbox{$\mathcal G$}}

\newcommand{\IC}{\mbox{$\mathcal I$}}

\newcommand{\PC}{\mbox{$\mathcal P$}}

\newcommand{\SC}{\mbox{$\mathcal S$}}

\newcommand{\RF}{\mbox{$\mathbb{R}$}}

\newcommand{\ZF}{\mbox{$\mathbb{Z}$}}

\newcommand{\ol}[1]{\overline{#1}}

\newcommand{\beq}{\begin{equation}}
\newcommand{\eeq}{\end{equation}}
\newcommand{\bear}{\begin{eqnarray}}
\newcommand{\bears}{\begin{eqnarray*}}
\newcommand{\eear}{\end{eqnarray}}
\newcommand{\eears}{\end{eqnarray*}}
\newcommand{\bdm}{\begin{displaymath}}
\newcommand{\edm}{\end{displaymath}}
\newcommand{\lba}{\left[\begin{array}}
\newcommand{\ear}{\end{array}\right]}

\newcommand{\degree}{\ensuremath{^\circ}}

\title{SymmGrid: Super-Scaling On-Robot Learning with Parallelized Symmetries and Egocentric–Exocentric Visual Perception.}
\author{
Gabe Everett, Brice Gunter, Ryan Vander Stelt,
Cleiver Ruiz-Martinez, Blake Hull and Juan Rojas\\
\thanks{EECE, Lipscomb University, Nashville, TN, USA}.
}
\begin{document}
\maketitle
\begin{abstract}
Deep reinforcement policy learning directly in physical robots (on-robot learning) remains bottlenecked by slow wall-clock training times. 
We present \textbf{SymmGrid}, a trajectory-level augmentation framework inspired by parallelized symmetries that super-scales group transformations to significantly accelerate on-robot learning in both egocentric and exocentric visual setups. 
We model a Markov Decision Process (MDP) under a symmetry tree, in which state--action pairs have admissible parallelized invariant transformations that yield a geometric grid structure. The state is modeled with ego- or exocentric images and proprioception information. 
The latter require special treatment, in the form of homographies, to warp visual scenes in line with their corresponding spatial transformations.
These parallelized transformations produce a large set of unique symmetric equivalences that populate the replay buffer with diverse and consistent experiences that speed up learning and improve performance.
We present extensive training and evaluations performed directly on real robot manipulation contact tasks including peg-insertions, cable routing, and object relocations.
Relative to SOTA, SymmGrid achieved wall-clock training convergence speedups of 1.37-2.17x, evaluation success rate improvements of 1.09x-1.27x, fastest training convergence times of 16.6, 10.9, and 79.3 minutes respectively. 
For trajectory wide assessments, we used normalized area under the curve (nAUC) ratios. SymmGrid achieved improvements of up to 2.59x. These results confirm that simple branch symmetries can have an outsized result due to super-scaling and bring us closer to sub-10 minute on-robot learning training in manipulation tasks suitable for arms and humanoids.
The project page is available at \url{https://symmgrid-robot.github.io/}.
\end{abstract}
\section{Introduction}\label{sec:intro}
\begin{figure*}[tbhp]
    \centering
    \begin{subfigure}{0.32\textwidth}
        \centering
        \includegraphics[width=2in, height=1.25in]{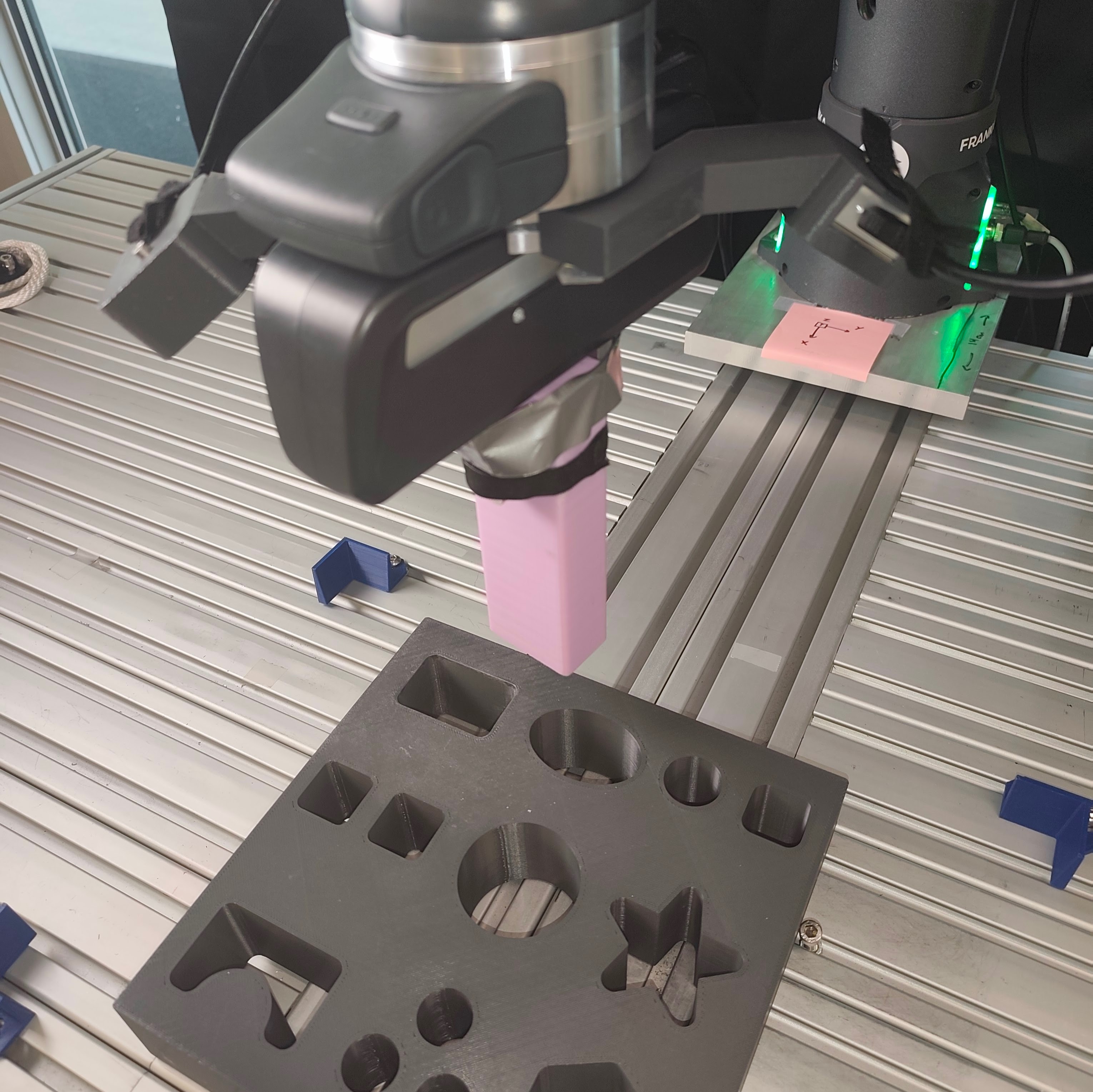}
        \caption{Peg-insertion.}
    \end{subfigure}
    \hfill
    \begin{subfigure}{0.32\textwidth}
        \centering
        \includegraphics[width=2in, height=1.25in]{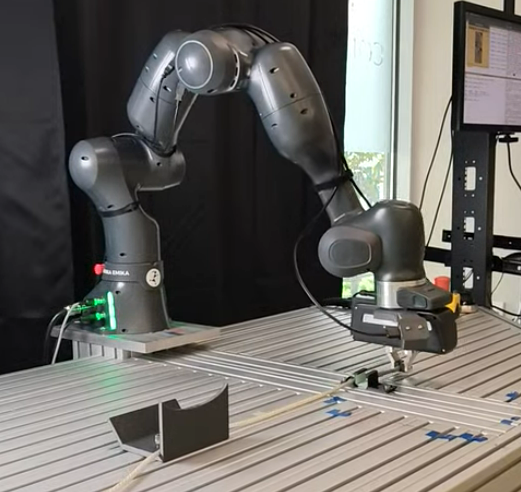}
        \caption{Cable-routing.}
    \end{subfigure}
    \hfill
    \begin{subfigure}{0.32\textwidth}
        \centering
        \includegraphics[width=2in, height=1.25in]{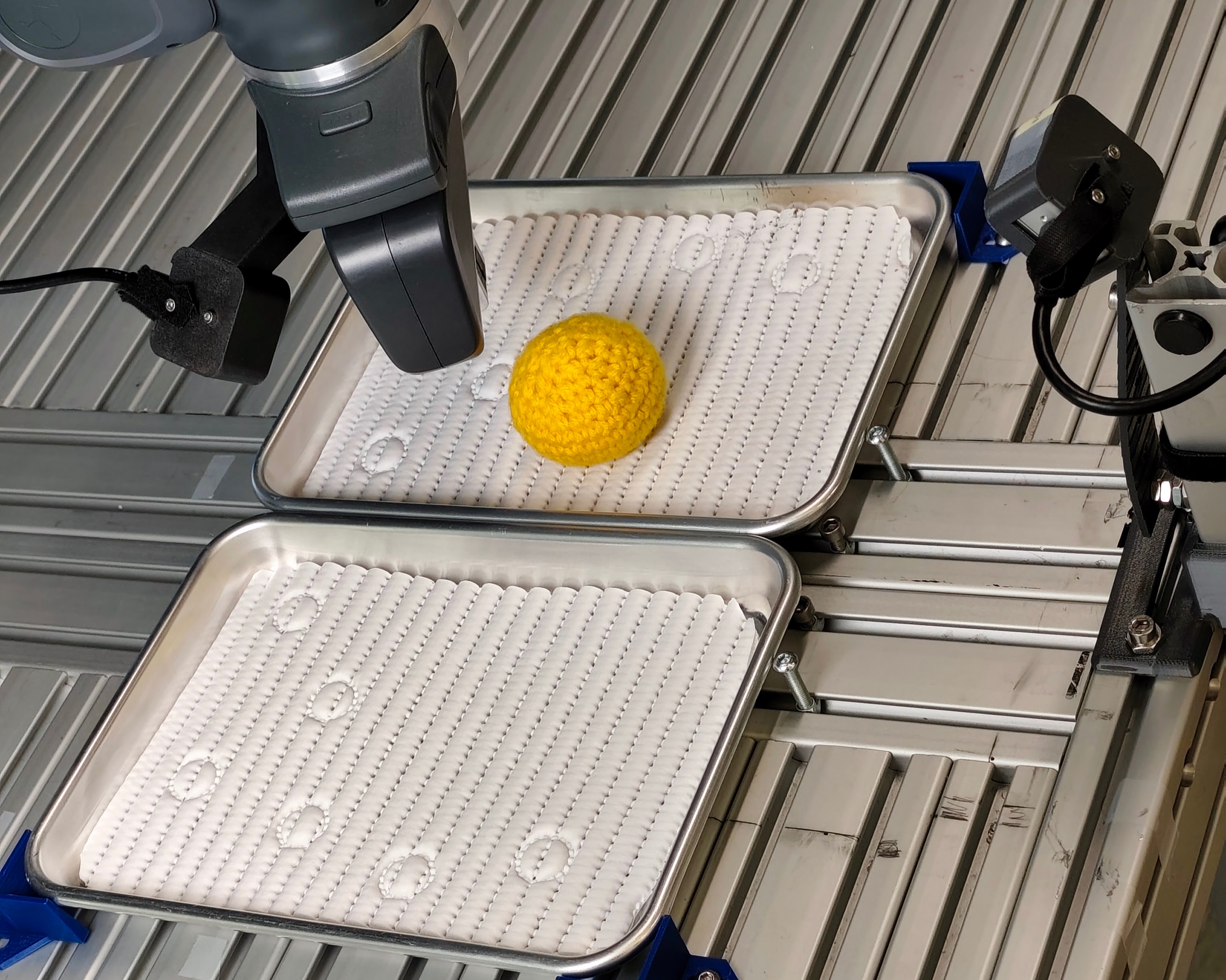}
        \caption{Object-relocation.}
    \end{subfigure}
    \caption{Real Franka research robot examples of manipulation contact tasks.}
    \label{fig:robot_setup}
\end{figure*}

Learning deep reinforcement learning (DRL) policies directly from physical robots (on-robot learning) is a desirable goal, as robots can learn world physics directly and avoid sim2real pitfalls. See Fig. \ref{fig:robot_setup} for a visualization of contact rich, on-robot learning tasks used in this work. Previous research has leveraged symmetrical principles and representations to accelerate learning and increase performance. Symmetry after all, underlies all physics \cite{Goldstein1950ClassicalMechanics}, and allows for invariant or equivariant representations.
Researchers have leveraged symmetry principles to augment data or build symmetry directly into the neural network architecture. SOTA on-robot learning currently train in about 20-180 minutes depending on tasks type and representation. These works typically bootstrap the algorithm using few demonstrations (about 20). The learning speeds achieved thus far are not practical for quickly deploying learning robots in the wild and promoting their wide adoption. A limitation is that the number of extracted symmetries from the original interaction remains significantly limited.

SymmGrid, presents two contributions:
(i) a method that leverages constant branched symmetries to accelerate policy learning in physical robots by super-scaling robot data generation. SymmGrid is conceptually simple, easy to implement, and significantly speeds up on-robot learning while improving performance, and 
(ii) a compute- and memory-efficient datastore process to handle branched symmetries with ego or exocentric visual setups. Exocentric visual setups leverage homographies at sample-time to handle visual warping and keep the transformed state invariant.

We conceive a Markov Decision Process (MDP) as a tree in which an episodic trajectory, composed of states and actions, is encoded by vertices and edges, respectively. An affine transformation can be applied to the trajectory to transform state-action (vertex-edge) pairs. These affine transformations can be applied such that a geometric structure, like a grid, is produced (see Fig. \ref{fig:branching_strategy_framework}a) and allows each vertex to hold a unique symmetric equivalence. 
In egocentric setups, visual scenes remain invariant with symmetric branching. As such, a pointer-based system is used in the replaly buffer to avoid copying identical images for the transformations. In exocentric setups, however, global fixed cameras cannot move in correspondence to the symmetric branching. As such, global images must be warped to match their expected views. We achieved this through efficient homography computations on the algorithm's sample-side. An index is appended to the replay buffer that indicates what kind of transformation that transition experienced and later used to elicit the correct homography transformation.
We present extensive training and evaluations results performed directly on real robot manipulation contact tasks including peg-insertions, cable routing, and object relocations.
Relative to SOTA, SymmGrid achieved point-wise wall-clock training convergence speedups between 1.37-2.17x, evaluation success rate improvements between 1.09x-1.27x, and fastest training convergence times of 16.6, 10.9, and 79.3 minutes respectively. For trajectory-wide results, SymmGrid achieved nAUC ratios of up to 2.59. These results confirm that simple branch symmetries have can have an outsized result due to super-scaling and bring us closer to sub-10 minute on-robot learning training.
\section{Related Literature}
We review contributions that speed up DRL policies via symmetry-based concepts in fully observable settings.  
Invariant Transform Experience Replay (ITER) \cite{2020RAL-Lin-ITER} transformed trajectory information $(s,a,r,s')$ via reflectional symmetry and trained a deep deterministic policy gradient. ITER produced up to 15 new reflectional symmetries on reach, slide, and pick-and-place simulation tasks and achieved a 3x-13x sample efficiency improvement relative to Hindsight Experience Replay (HER) baseline \cite{2017NeurIPS-Andrchowicz-HER}.
%
Wang \et embedded a cyclic group of dimension 12 $C_{12}$ to encode rotational symmetry directly in a Fully Connected Network (FCN) \cite{2021CORL-Wang-EquivQLearn_SpatialAction}. The FCN was convolved with an image, producing an action heatmap to instantiate picks. The system produced 3x sample efficiency improvements against RAD \cite{2020NeurIPS-Laskin-RAD} and DrQ \cite{2020ICLR-Ilya-DrQ}.
%
In \cite{2022ICLR-Wang-SO2EquivRL}, SO(2)-EquivariantRL defined an equivariant Q-function and an invariant optimal policy, leading to the production of Equi-DQN and Equi-SAC, respectively. These used depth images and displacements and were tested with a number of algorithms, achieving significant performance gains and around a 10x improvement in sample efficiency.
%
In \cite{2022CoRL-Wang-OnRobotLearn}, SO(2)-Equivariant RL learned online via equivariant learning. In \cite{2022ICLR-Wang-SO2EquivRL}, Equi-SAC used a dihedral group $D_4$ in the NN and an augmentation of four random SO(2) samples to train across block picking, clutter grasping, block pushing, and block-in-bowl. In this work, a UR5 robot was trained between 45 minutes and 2 hours 40 minutes of wall-clock time for meaningful performance.

The presented methods use only a few symmetries in their data generation process. We explore the use of branched symmetries to \textit{super-scale} the generation of possible symmetries and their effect on improving wall-clock training times, sample efficiency, performance, and robustness. 
In \cite{2026Frontiers-Vanderstelt-FractalSERL}, a \textit{brief research report on preliminary work} outlined a broader architecture for accelerating on-robot learning with fractals and primarily demonstrated the proof of concept in simulation. SymmGrid focuses on a special case of fractal symmetries: constant branching. In this paper, we presents extensive and detailed on-robot training and evaluations results, including visual scene handling for ego- and exocentric setups, yielding significant efficiency and performance results relative to SOTA.
%
\section{Preliminaries}
\subsection{Branching Systems}\label{subsec:prelim_fractal_sys}
Branched symmetries pervade nature and use simple rules to build efficient structures. We draw inspiration from the Racemose plant (especially the Corymb Inflorescence variety), which branches radially alongside the stem to produce a flat-topped or convex crown \cite{MasterGardenersofNorthernVirgina2026CorymbInflorescences}.

Branching systems can be modeled via Iterated Function Systems (IFS) \cite{1981-hutchinson-fractals}. Consider a tree $\mathcal{G}$ with vertices $V$, edges $E$ and depth $h$ such that $\GC=(V,E)$. The tree has spatial geometric properties and branching dynamics we can model.

Consider a vertex $v_k$, which describes a pose of interest $\xb \in \RF^n$ at a depth $h_j$. Vertices $v_k$ are equivalent to branches and can be produced via an Affine transformation $\TB_k(x) = \AB x + t_k$. $\TB_k(x)$ is an $n$-dimensional Euclidean group $\EB(n)\in(\AB,t)$ transformation, where $\AB \in O(n)$ is a member of the Orthogonal group and $t \in \RF^n$ translates $x$. Vertices can be arranged to form a desirable structure: linear branches, planar $K \times K$ grids, or more complex geometries. Fig \ref{fig:branching_strategy_framework}a shows an example of a grid structure. 

A branching rule $b$ determines how each vertex branches when activated. In this work, we adopt a constant branching strategy $b_{const}(0)$ that yields parallelized symmetries and inputs a proportional distribution of transformations into the replay buffer.
\subsection{Markov Decision Processes}\label{subsec:mdp}
We model an MDP via the tuple $\langle \SC, \AC, \PC, R, p, \gamma \rangle$, where $\SC$ is a continuous state space; $\AC$ is a continuous action space; $\PC: \SC \times \AC \times \SC \to [0, 1]$ is the unknown transition function that describes the environmental dynamics $p(s'|s, a)$; $R(s, a, s')$ is the reward when an agent reaches state $s' \in \SC$ after performing action $a \in \AC$ in state $s \in \SC$ and can be dense or sparse; $p(s_0)$ is the probability distribution over initial states and $\gamma \in [0, 1]$ is a discount factor. The continuous robot learning problem corresponds to the RL problem of obtaining a parameterized stochastic policy $\pi_{\theta}: \SC \to P(\AC)$ that maps states to a distribution of actions, such that the expected maximum-entropy objective is maximized for any given goal \cite{Haarnoja2018SoftActor}. 
\subsection{SERL}\label{subsec:serl}
Sample Efficient Reinforcement Learning (SERL) \cite{2023CORL-Luo-SERL} is a software suite for sample-efficient on-robot reinforcement learning. It is composed of sample-efficient off-policy DRL methods; a variety of reward generation methods (engineered and sparse image-based binary classifiers); environment reset methodologies for reset-free physical training; a safe RL high-quality impedance controller, and four contact tasks (peg-insertion, PCB assembly, cable routing, and object-relocation). SERL trained policies with SOTA wall-clock times for on-robot learning averaging between 36-105 minutes. 

For its most efficient variant, SERL uses Reinforcement Learning with Prior Data (RLPD) \cite{2023ICML-Ball-RLPD} (which builds on DrQ \cite{2020ICLR-Ilya-DrQ}), uses both demo and online data across two separate replay buffers, and runs the actor and the learner(s) asynchronously. All tasks use an impedance controller running at 1K Hz. Peg-insertion and PCB assembly use a position-based norm to produce a sparse reward. Cable-routing and object relocation use an image-based binary classifier to produce sparse rewards set as $r(s) = \mathbf{1}\left[p(e \mid s) \geq 0.5\right]$. The classifier is trained with both positive and negative examples. A well trained classifier is crucial for consistent rewarding and learning.
Note that the RLPD does not implement off-the-shelf light randomization. As such, RLPD is sensitive to light changes and requires consistent lighting for successful training and deployment.
SymmGrid will compare performance against SERL and will use it as a baseline for performance evaluations.
\section{Branched Markov Decision Processes}\label{sec:fractal_mdp}
Branched MDPs can super-scale robot learning by leveraging invariant transformations in RL rollouts. 
Consider an episodic RL setup with a trajectory $\tau$ of finite duration $t_{max}$ as sequences of state, action, reward tuples: 
$\lip s_0$, $a_1$, $r_1$, $s_1$, $a_2$, $r_2$, $s_2, \ldots, s_{t_{max}} \rip$. 
For a branched MDP, each vertex represents a state $s \in \SC$ and each edge represents an action $a \in \AC$. 
We define the set of all executed (contact-rich) trajectories as
$\ol \Gamma = \cup_{h=1}^H \SC \times (\AC \times R \times \SC)^h$. A global affine symmetrical transformation $\sigma$ applied to each contact-rich feasible trajectory is an invariant one-to-one mapping $\sigma:\ol \Gamma \rightarrow \ol \Gamma$ that generates a newly formed physical rollout:
$\sigma(\ol \Gamma) = \{ \tau' \in \ol \Gamma \mid \exists \tau \in \ol \Gamma, \sigma(\tau) = \tau' \}$ \cite{2020RAL-Lin-ITER}. 
The affine transformation can be understood as a decomposable symmetry that applies to the trajectory's states and actions: 
$\sigma_{\SC} : \SC \to \SC$, and $\sigma_{\AC} : \AC \to \AC$ and are reward-preserving. The rewards (and returns) in $\tau'$ appear exactly as in $\tau$. 
The trajectory tuple becomes: $\tau' = \lip s_0', a_1', r_1', s_1', a_2', r_2', s_2', \ldots, s_{t_{max}}'\rip$, where the symmetry mapping is applied to each MDP element at every time step in the episode:
$s'_0 = \sigma_{\SC}(s_0)$, 
$a'_i = \sigma_{\AC}(a_i)$, 
$s'_i = \sigma_{\SC}(s_i)$, and
$r'_i = R( \sigma_{\SC}(s_{i-1}), \sigma_{\AC}(a_i), \sigma_{\SC}(s_i) )$.

In SymmGrid, the symmetry operator can be set to a reward-invariant and dynamics-invariant branched symmetry operation: $\sigma(\cdot): \TB_k(\cdot) = \IB(\cdot) + t_k$. 
The transformation is applied to both states (vertices) $s_i' = \sigma_{\SC}(s_i):\TB_k(s_i) = \IB(s_i) + t_k$ and actions (edges) $a_i' = \sigma_{\AC}(a_i):\TB_k(a_i) = \IB(a_i) + t_k$. 
The transformations are conducted to produce a desirable geometric branched structure (line, grid, or more complex shapes introduced in Sec. \ref{subsec:prelim_fractal_sys}) as shown in Fig. \ref{fig:branching_strategy_framework}a. The result is a finite set of transformed vertices and edges (including the original pair) at each time-step. To clarify, branching transformations occur at tree depth $h$, which may differ from MDP time steps, depending on the branching factorization strategy one pursues. In SymmGrid, with constant branching, they are always equivalent. 
\subsection{Branching Geometric Structure and Affine Global Transformations}\label{subsec:branch_geom_struc_affine_global_trans}
\begin{figure}[tp]
    \centering
    \begin{subfigure}[c]{0.2\textwidth}
            \includegraphics[width=\linewidth]{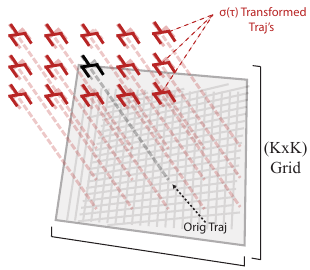}
            \caption{Visualization of symmetric branching of a trajectory for a $K\times K$ grid.}
    \end{subfigure}
    \hfill
    \begin{subfigure}[c]{0.24\textwidth}
        \centering
        \includegraphics[width=\linewidth]{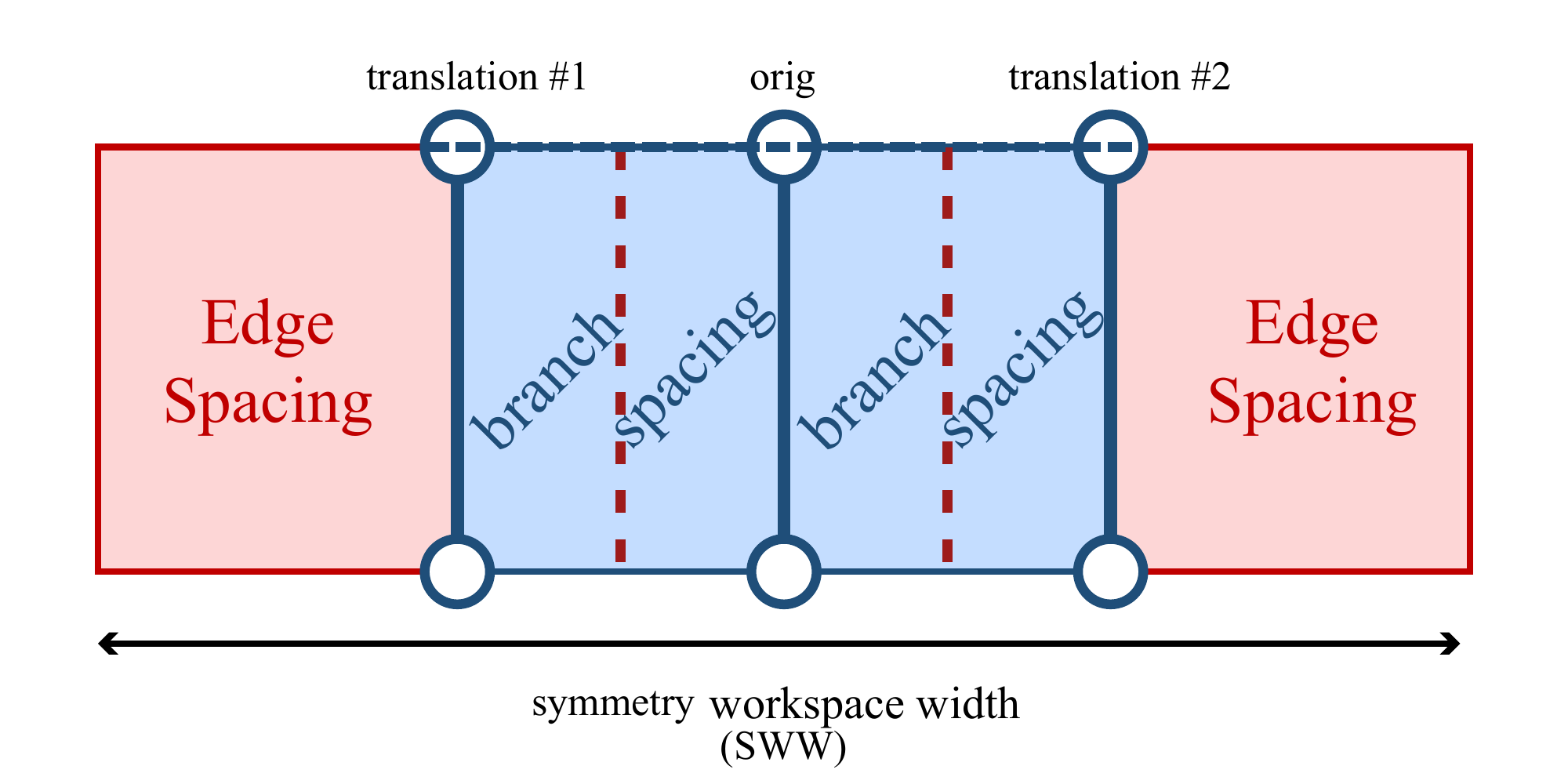}
        \caption{Cross-section grid spacing for 3 branches (inclusive).}
    \end{subfigure}
    \caption{Constant branching visualization.}
    \label{fig:branching_strategy_framework}
\end{figure}

We select affine transformations to produce a $K \times K$ grid geometric structure conformed by vertices that include the original trajectory, where $K \in \ZF$. A grid visualization is shown in Fig. \ref{fig:branching_strategy_framework}a. Note that the structure is generated immediately when the MDP starts at depth $h_0$ in the tree \cite{2026Frontiers-Vanderstelt-FractalSERL}. 


Recall that a trajectory symmetry is decomposed into state and action transformations: $\sigma(\ol \Gamma):\sigma_{\SC}(\cdot) \bigcup \sigma_{\AC}(\cdot)$. Care must be taken in the way that symmetries are applied, in particular to states. In our formulation, transformations are global rather than local. State representations may include proprioception (pose, twist, wrench, and finger) information along with more complex representations (images, point clouds, etc). In SymmGrid, we limit the problem to RGB images $\IC$ and proprioception information. Note that egocentric images have unchanged scenes after transformations ($\sigma_t(\IC)=\IC$) while exocentric images do not.


The motivation behind the combinatorial set is to aid in super-scaling the robot interaction. Consider a $27 \times 27$ grid. A total of 729 vertices (including the original one) are spawned for each unique symmetrical transformation at a given time-step.
The entire generated data set is inserted into the replay buffer $\BC$ before sampling and optimization: $\BC \lar \sigma(\ol \Gamma)$. 
When images are involved, our implementation instantiates a separate image buffer. 
Given that we use global transformations, the agent and the objects are transformed identically. The images should be unchanged\footnote{A larger discussion about background scenes and assumptions will be discussed in limitations.}.
Handling images in the replay buffer can be a compute intensive hand. For unchanged scenes, we introduce pointers into the replay buffer to ensure we do not enter repeated images. For exocentric scenes that change, an efficient routine that leverages homographies is used (see Sec. \ref{sec:homographies}).

To optimize wall-clock efficiency, all transformations in the $K \times K$ grid are computed in parallel. We present computations for the grid translations $t_{(k,k)} \in K \times K$ along the robot's base $xy$-plane.
We define a square ``symmetry workspace'' (SymWS) within which all transformations must lie. The SymWS is constrained to exist within the larger robot workspace. The SymWS is always centered at the robot's end-effector position. 
The edge corner location of the SymWS is computed by subtracting the robot's end-effector $(x,y)$ position from the SymWS half-width: $SymWS_{(0,0)}= s_{(x,y,\cdot)} - \frac{SymWS_{width}}{2}$.
Given this knowledge, transformations are measured from the edge of the SymWS, not from the end-effector's center location. At each end of the grid, there will always be some degree of free space, denoted as "Edge Spacing" in Fig. \ref{fig:branching_strategy_framework}b in red. This edge spacing is a function of the number of branches (inclusive of the original) at that depth (3 branches in our example in Fig. \ref{fig:branching_strategy_framework}b). The branch positioning is controlled by: $t_{{k_x,k_y}}$, where each $k$ is an index into the grid on the $xy$-plane; the branch index $b_{i,j}$, which starts at 1, and the total branch count for that depth $b_{c_h}$. Iterating from one grid corner to the other:
\begin{equation}
    \centering
    t_{{k_x,k_y}}=SymWS_{(0,0)}+\frac{(2*b_{i,j}-1)*SymWS}{2*b_{c_h}}.
    \label{eqtn:grid_translations}
\end{equation} 
\subsubsection{Measuring Performance}\label{subsubsec:generalized_metric}
In DRL, when measuring learning speed-ups in robot learning, pointwise metrics are typically used. For example, the number of samples to achieve a particular return, or the maximum performance in the task. We believe trajectory-wide metrics are particularly important in measuring the overall efficiency and robustness performance of a policy. For this reason, we use nAUCs and the ratio across methods. Note that nAUC values closer to 1.0 tend to have return or success ratio curves that tend to the upper-left corner. The larger the area of one metric over another, the better the overall efficiency and performance. Once nAUCs have been computed, we can compute the ratio across methods to get a relative idea of performance and efficiency gains across the policy. 
\section{Homographies and Invariance}\label{sec:homographies}
When an invariant transformation is applied to the MDP, states and actions are transformed accordingly. Egocentric images remain invariant as their perspective is unchanged. Exocentric images on the other hand, experience perspective changes, and as such require an efficient process by which the original image is modified to approximate such changes. To this end, we leverage homographies. 

A homography $H$ is a projective mapping of image points from one \textit{plane} representation to another \cite{homography_opencv} according to Eqtn. \ref{eqtn:homography}:
\begin{equation}
    H(\Delta) = MT(\Delta)M^-1
    \label{eqtn:homography}
\end{equation}
It leverages a calibration Matrix $M$ that can map planar homogeneous Cartesian coordinates to homogeneous image pixel coordinates according to Eqtn. \ref{eqtn:homography_calibration} and can be computed with OpenCV's \verb|findHomography()| method \cite{homography_opencv}.
\begin{equation}
    \bar{q} = M\bar{p}.
    \label{eqtn:homography_calibration}
\end{equation}
It also uses a matrix of planar transformations. For SymmGrid, we consider planar translations of the form $\Delta=(\Delta x, \Delta y)$:
\[
T(\Delta) =
\begin{bmatrix}
1 & 0 & \Delta x \\
0 &  1 & \Delta y \\
0 &  0 &  1
\end{bmatrix}.
\]
Eqtn. \ref{eqtn:homography}, achieves the projective mapping by taking the original image scene, converting it a point in physical space, transforming it, and then re-projecting the translated point back into the image. In SymmGrid, given that the transformation information is available \textit{a priori} (see Eqtn. \ref{eqtn:grid_translations}), a matrix of custom homographies $H_b$ can also be computed \textit{a priori}:
$
    H_b(\Delta) = MT_b(\Delta)M^-1
$
Note that the mapping depends on image coordinates. As such, one must calibrate using an image of equal resolution to the learning system. A scaling matrix can be used to perform the calibration with a larger image resolution before scaling the computations down. 
The calibration must use the same reference coordinate frame as the learning system. This is relevant to SymmGrid, as the learning system uses the end-effector frame via a wrapper to improve neural network optimization.
Also, in homography, instead of working directly with the forward mapping where, for a given input pixel, one computes the equivalent output pixel, a backward mapping is used. The latter traverses the output pixels and identifies what input pixel maps to it to avoid leaving holes in the output image. Note however that not all output pixels have a matching input pixel. Some output mappings originate in pixels outside of the original image. In these cases, filling strategies are used. Edge pixel replication appears to be the most advantageous strategy for facilitating neural network training, particularly when compared to uniform solid-color padding approaches.

Furthermore, to improve computational efficiency we make use of two insights. First, note that \verb|findHomography()| computes a reverse and forward map in each call. In constant branching, where transformations will not change, reverse mapping can be computed from the start. Second, instead of computing the homographic transformation to the original global image before inserting into the replay buffer and holding large number of modified images, we can use \verb|remap()| during sampling time, and only modify the original state egocentric image $K^2$ number of times. This approach leverages our efficient image handling process presented in Sec. \ref{subsec:branch_geom_struc_affine_global_trans}. 
To identify the correct transformation, each transformation is identified with an index value, which is also entered into the buffer. At sample time, the index value is used to extract the corresponding transformation, which is computed in place.
\section{Experiments and Results}
We present experimental results for peg insertion, cable routing, and object-relocation using SymmGrid for on-robot learning as shown in Fig. \ref{fig:robot_setup}. We conduct extensive training and evaluations. 
All policies leverage a pre-trained ResNet-10 as a vision backbone (it later connects to robot proprioception information and is processed via a 2-layer MLP for the policy and Q-network). The policy outputs a 6D or 7D end-effector delta pose, which is executed by a low-level impedance controller. Note that angular action updates are capped at $\pm$0.01 rad for roll and pitch and $\pm \pi/6$ for the yaw axis. This essentially only allows for rotations about the z-axis. 
Currently, all tasks collect 20 demonstrations to bootstrap the RLPD algorithm except for object relocation which used 30.

As for state transformations, we limit ourselves to pure translations $\sigma_t(s)$ and only affect positional components $\sigma_{\tb_k} but not orientation, twists, wrenches, or fingers: (s)=s[\IC, \xb_{eff} + \tb_k,\theta,\vb,\wb,g]$. The other state components remain unchanged under translation. As for the action space, the transformation acts trivially on the action $\sigma_{\tb_k}(a)=a$. The same control inputs are executed, but from translated states. In summary, $\sigma_{\tb_k}(\tau)= (\sigma_{\tb_k}(s),a)$. 

Three metrics are used to report results: efficiency (wall-clock time), performance (success rate), and nAUC (and its comparative ratio). The first two are evaluated at specific points, and the last provides a trajectory-wide comparison.
Our evaluations measure five seeds of on-robot learning (three for object relocation) at multiple step intervals for both training and evaluation—a significant effort for direct evaluation on physical robots. 

Training results are reported as success rates averaged over 50 samples. Evaluation results are presented as the success rate across 50 task attempts. Unlike others, our results enable us to closely track the policy's mean learning dynamics across for training and evaluation. 
Often, on-robot learning works only report a single success rate for their best-trained policy due to the cost of executing these experiments.

Preliminary research in \cite{VanderStelt2026ExploringSymmetries} indicated that a $27 \times 27$ grid size, a 0.3 m symmetry workspace width, and a 3.6 million entry replay buffer are optimal We use these values for peg insert and object relocation and indicate differences for object relocation where appropriate. 
SymmGrid uses the same set of hyperparameter settings as SERL (except for SymmGrid-specific hyperparameters) to maintain equivalent comparisons \cite{2023CORL-Luo-SERL}. The full listing is shown in \cite{VanderStelt2026ExploringSymmetries}. Note however that SymmGrid used an AMD Ryzen Threadripper 1950x processor with 16 cores/32 threads running at 3,400 MHz, 128 GB RAM, and an NVIDIA RTX 4070 GPU with 12 GB of VRAM. SERL's original results \cite{2023CORL-Luo-SERL} used an RTX 4090, which has \textbf{2.84x} the FP32 TFLOP capacity (~82.6 vs 29.1) and \textbf{2x} the bandwidth capacity (1008 GB/s vs 504 GB/s), making our results all the more significant.
\subsection{Peg-Insertion Experiments}\label{subsec:peg-insert-exps}
The FR3 peg-insert task was originally codified in \cite{Luo2024FMB:Learning}. Here, a robot inserts a specific-shaped peg into its corresponding slot in a board. 
The state space includes six components: $s[\IC, \xb_{eff},\theta,\vb,\wb,g,\xb_{targ}]$, two (128 x 128) images from wrist-mounted cameras, the end-effector's position and orientation, linear and angular velocities, and the gripper position.

The action space is a 7-dimensional space that enacts deltas for position (3), orientation (3), and gripper (1) commands  $a[\Delta \xb, \Delta \theta, \Delta g]$. All actions are constrained within a predefined bounding box to ensure safe execution.
We use a sparse reward and terminate episodes upon task completion or 100 steps.
SymmGrid to uses the same impedance controller and relative task frame as SERL.
We bootstrap the policy with 20 demonstrations and sample 50\% from each of the prior and online data replay buffers.
We used \textit{random resets} at the start of each episode and terminated after 100 steps.
\subsubsection{Peg-Insertion Results}
\begin{figure*}[htbp]
    \centering
    \begin{subfigure}[t]{0.49\textwidth}
        \centering
            \includegraphics[width=\linewidth]{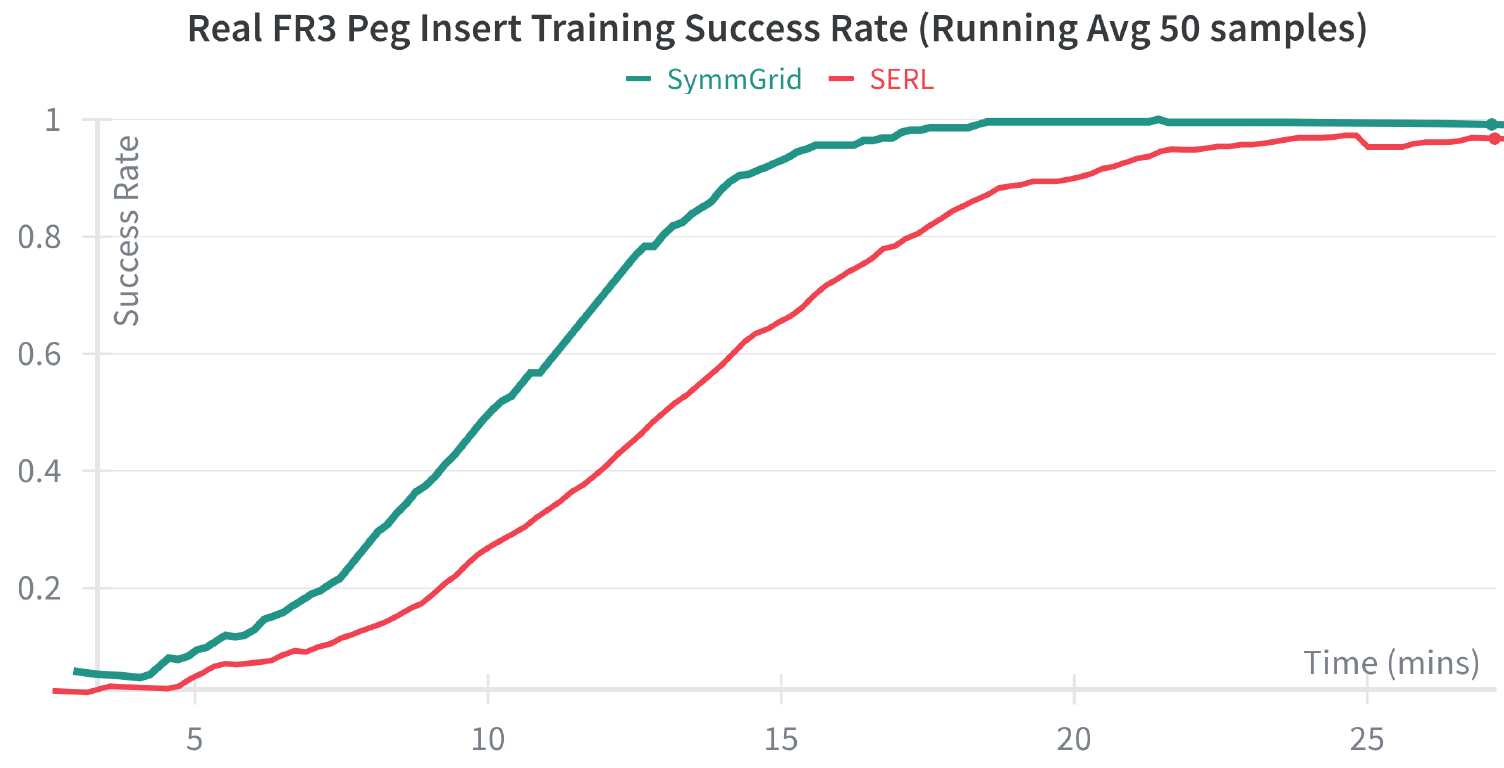}
            \caption{Real training success rate vs. wall-clock time. The plot was smoothed by a 50-sample running average. SymmGrid converges to 100\% success rate in 18.5 mins \vs 26 mins. A 40.5\% speedup in training convergence.}
    \end{subfigure}    
    \hfill
    \begin{subfigure}[t]{0.49\textwidth}
        \centering
            \includegraphics[width=\linewidth]{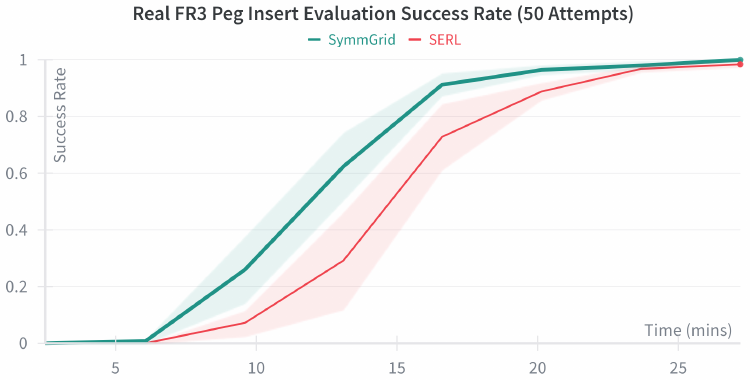}
            \caption{Real evaluation success rate vs. wall-clock time. 50 insertions attempted. SymmGrid achieves an $\sim90\%$ success rate at around 15.9 mins, while SERL at 20.13 mins, a 27\% speedup.}
    \end{subfigure}    
  \caption{SymmGrid \vs SERL peg-insertion results for on-robot learning. Trained across five seeds using a constant grid size of $K=27$. The envelope depicts the standard error of the mean. }
  \label{fig:peg_insertion_results}
\end{figure*}

Peg-Insertion results for the real Franka robot with SymmGrid are shown in Fig. \ref{fig:peg_insertion_results}. 
As for pointwise evaluation, SymmGrid's training converged to 100\% accuracy in as little as 14.77 minutes and, on average, 18.5 minutes, while SERL converged at 26 minutes on average. SymmGrid on average trains 1.41x faster and reduces training time by 28.8\%.

Evaluations were conducted every 500 steps. SymmGrid achieved a 100\% success rate as fast as 16.6 minutes and an average 90\% success rate in 15.9 minutes, while SERL took 20.13 minutes, resulting in a 26.6\% speedup. Similarly, SymmGrid achieved a 96.4\% success rate at 20.13 minutes, while SERL took 22.7 minutes\footnote{Note that the original work reported peg-insert with random-resets takes 60 minutes to converge to a 100\% success rate it is not clear why our baseline performed better than the cited work (see point 9 \cite{github-serl}).}, that is a 1.28x speedup (or a 2.64x speed up considering the original source). SymmGrid also reached a 100\% success rate at 27 minutes. 

The trajectory-wide nAUC compares the algorithms' overall performance. For peg-insertion, we compute the nAUC up to a 3500-step marker. SymmGrid's nAUC is 0.607 whilst SERL's is 0.499, a 1.22x increase. 
\subsection{Cable-Routing Experiments}\label{subsec:cable-route-exps}
The FR3 cable-routing task requires the robot to insert a rope into a bracket on the table. The robot starts in random positions and orientations within the workspace, and the rope is pulled by a lightweight nut to keep it snug. This is a challenging task for model-based control under rigid-object assumptions, given the rope's deformable nature. 
The state space and action space are identical to the peg-insert setup.
For this task, we use a sparse reward and the task terminates upon completing 100 steps.
\subsubsection{Cable-Routing Results}
\begin{figure*}[htbp]
    \centering
    \begin{subfigure}[t]{0.49\textwidth}
        \centering
            \includegraphics[width=\linewidth]{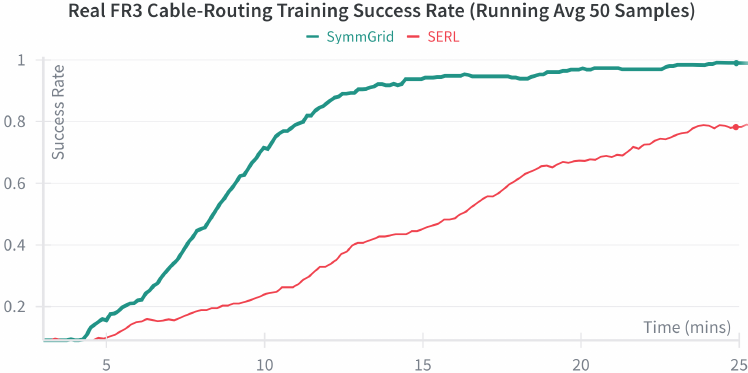}
            \caption{Real training success rate vs. wall-clock time. SERL reaches a 78.9\% success rate in 23.9 mins. SymmGrid achieved the same result in 11.0 mins. A 117.3\% speed up in training.}
    \end{subfigure}    
    \hfill
    \begin{subfigure}[t]{0.49\textwidth}
        \centering
            \includegraphics[width=\linewidth]{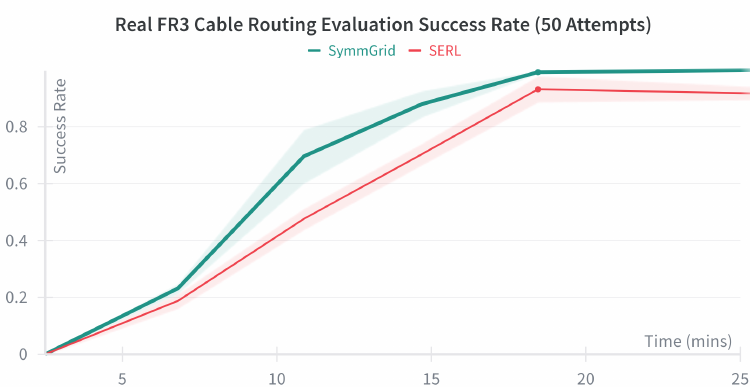}
            \caption{Real evaluation success rate vs. wall-clock time. 50 insertions attempted. SymmGrid achieves a 99.2\% success rate at 18.5 mins, while SERL does not reach this metric within 25 mins.}
    \end{subfigure}    
  \caption{SymmGrid \vs SERL cable-routing results for on-robot learning. Trained across five seeds using a constant grid size of $K=27$. The envelope depicts the standard error of the mean. }
  \label{fig:cable_routing_results}
\end{figure*}

Cable-routing results for the real Franka robot with SymmGrid are shown in Fig. \ref{fig:cable_routing_results}. 
As for pointwise evaluation, SymmGrid's training converged to 100\% accuracy in as little as 11.9 minutes and on average 98\% in 22.8 minutes, while SERL did not reach the latter within the 25 minutes considered for the task; it only reached a maximum average accuracy of 74.8\%.

Evaluations were conducted every 1000 steps. SymmGrid achieved a 98\% success rate as fast as 10.9 minutes and a 90\% success rate in 15.9 minutes, while SERL took 20.13 minutes, resulting in a 1.27x speedup or 21\% time reduction. Similarly, SymmGrid achieved a 93.2\% success rate in 17 minutes, while SERL took 18.5 minutes, resulting in an 1.09x speedup. SymmGrid also achieved a 99.2\% success rate at 18.5 minutes, while SERL did not achieve this within the 25-minute allotted time. 

The trajectory-wide nAUC for cable-routing was computed using 6000 steps. SymmGrid's nAUC is 0.63 whilst SERL's is 0.58, a 9.0\% increase. 
These evaluation results surprised us given the large spread seen during training time. Perhaps more seeds or further analysis is needed to better understand these numbers.
\subsection{Object-Relocation Experiments}\label{subsec:obj-reloc-exps}
The object-relocation task is setup with two small trays placed side-by-side at the center of the workspace, with a soft compressible ball placed in the starting tray for the forward model. A reset free learning system is implemented by using both forward and backward models. The image-based binary reward classifier is used for both such models. A small negative living reward is also used every time the grippers move to encourage conservative pick attempts. 

With regard to perception, this task differs significantly from the other two in a number of ways: (i) it combines ego- and exocentric view with one rear wrist-mounted camera and a front facing global camera; (ii) the task is partitioned into forward and backward policies to avoid a reset-free training. We used the standard Franka FR3 parallel jaw gripper for this task. To maximize visibility, which is crucial for learning, the gripper was rotated by $\pi/2$ along the global z-axis, leaving the gripper's base aligned with the global y-axis to expose a direct view of the gripping to the front cameras as seen in Fig. \ref{fig:robot_setup}c.  The front-facing wrist-mounted camera used in the other tasks was also removed to reduce occlusion, especially when the end-effector works near the front camera. A total of 30 demonstrations were used to train the policy. This higher amounts was selected before visibility problems were identified as a culprit in early failed training attempts. The number was retained and does not affect baseline comparisons as SERL use the same quantity.

As for the homography, the calibration homography matrix $M$ used 20 non-collinear points using a fixed height set at the base of the trays. The points spanned the space of both trays. The output calibration matrix $\text{M}_{\text{target}}$, which undergoes pixel-centered resizing from the original 640 x 480 image to a 128 x 128 resolution that matches SymmGrid's was: 
\[
\text{M}_{\text{target}} =
\begin{bmatrix}
-66.3742 & 99.7526 & 62.9547 \\
  5.5959 &  4.3313 & 21.4402 \\
 -1.0581 &  0.0157 &  1
\end{bmatrix}.
\]
The produced calibration matrix was used along with $b=K^2$ shifts 
for custom homographies and an end-effector coordinate frame.
A varying number of branches and symmetric workspace widths were tested across three seeds as well as the enabling of homographies. Note that in contrast to previous works, this research presents (the very expensive) but full training curve dynamics for real robot training across seeds and ablations. Experimental results are 
in Sec. \ref{subsubsec:obj_rel_results}. Object relocation was tested in SymmGrid using constant branch values of 27, 9, and 3 constant branches as well as symmetric workspaces of 0.30 m and 0.175 m.
\subsubsection{Object-Relocation Results}\label{subsubsec:obj_rel_results}
In this section, we present homography results, highest performing ablations, and qualitative descriptions.

\begin{figure}[bt]
    \centering
        \includegraphics[width=0.65\linewidth]{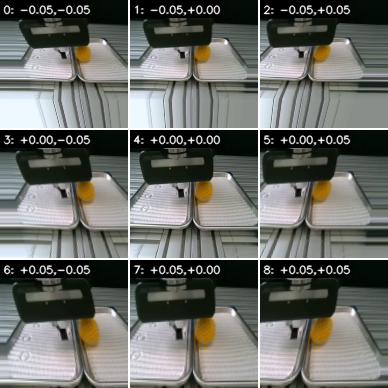}
        \caption{Visualization of homography shifts in relation to end-effector relative frame. Shifts are displayed atop of each figure in meters with the central figure being the original image. Notice the artifact formation in certain scenes.}
    \label{fig:homography_shift}
\end{figure}

We visualize homography results for 3 constant branches in Fig. \ref{fig:homography_shift}. Note that during batching, random transitions are sampled from the buffer. If we were to visualize those images, a randomized set of shifts would be seen. For clarity, we display a set of 9 clean shifts relative to the end-effector frame.
We also present calibration error tests. In our homography calibration, 20 pixel-robot-position correspondences were used under a 640 x 480 image from the front camera (before scaling). 16 points were admitted as RANSAC inliers with a root-mean-square error (RMSE) error less than 3 px. The inlier RMSE was 1.387 px with a maximum inlier error of 2.240 px. In context, the RMSE inlier error constitutes a very small residual error that produces consistent shifts in transformed images.

\begin{figure}[bh]
    \centering
        \includegraphics[width=\linewidth]{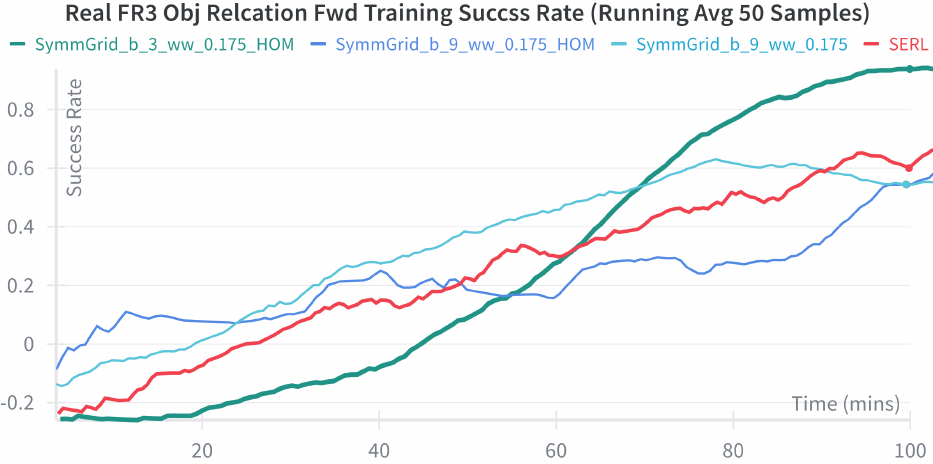}     
        \caption{Object Relocation Forward Training Results: 3 SymmGrid ablations presented relative to SERL using the more narrow workspace width of 0.175 m across 3 seeds.}
        \label{fig:object_relocation_results}
\end{figure}

We now present our highest performing ablations for training results. Particularly 9 and 3 constant branches and a symmetric workspace width of 0.175 m.
Fig. \ref{fig:object_relocation_results} shows the top performing ablations vs. the baseline SERL for the training of object relocation for the forward model.
As for pointwise evaluation in wall-clock training time (Fig. \ref{fig:object_relocation_results}a), SymmGrid's training converged to 92\% accuracy in as little as 79.3 minutes and on average in 94.6 minutes, while SERL reached the same value on average in 130.0 minutes, a 35.4 minute difference or 37.4\% speedup\footnote{Note that relative to the original SERL experiment, the more challenging object used here delays training.}. At 100 mins, SymmGrid had reached an average success rate of 93.7\% \vs SERL's 60.0\%, a difference of 33.7\%. For sample step performance, the difference is more dramatic. SERL achieves a 60\% success rate at 10,000 steps, while SymmGrid achieves it at 5851 steps, a 70\% improvement. 


As for point-wise evaluations, given this task's extensive duration, only a single checkpoint, at 10,000 steps, was used for evaluation across the three seeds. The baseline succeeded 74.7\% for the forward model and 79.3\% for the backward model. SymmGrid, on the other hand, achieved 99.3\% and 95.3\% success relocations respectively. Thus, SymmGrid achieved accuracy gains of 33.0\% and 20.2\% respectively.
For trajectory-wide evaluations, SERL got a nAUC of 0.1911, while SymmGrid got a nAUC of 0.494, a 2.59 ratio.

\begin{figure}[bt]
    \centering
        \includegraphics[width=\linewidth]{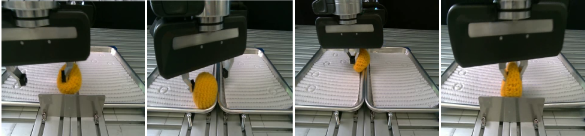}
        \caption{Visualization of emergent behaviors in moving a soft ball across trays: external finger push, inter finger drag, pinched finger push, and regular pinch.}
        \label{fig:emerging_behs}
\end{figure}

Finally, we describe some qualitative results. When we first setup the task, we naively chose a soft crocheted ball (contrast with a sturdier flat strawberry selected in the baseline experiments). Such ball has unique dynamics: it can easily roll, slide, and be squashed. Contrary to the unique pinching behavior seen in SERL \cite{2023CORL-Luo-SERL}, in this work we saw the robot elicit a number of emerging behaviors including: sliding the ball with both the interior or exterior parts of the finger, pressing and dragging, pinching, and combinations of these. Fig. \ref{fig:emerging_behs} shows snapshots of these.
The system also worked proved resilient to different disturbances. When either camera was blocked, the end-effector would balance in place, but resume once the occlusion was over. If the ball was displaced both before it reached out or even during a pinch, the end-effector would reactively seek the ball's new location and attempt to get it over. The system and the classifier also performed fairly consistent when one or both baking trays were removed or slanted up to a nominal amount (\ie 30\degree{} of the vertical).
Results are summarized in Table \ref{tab:eval_comparison}
\begin{table}[t]
\centering
\caption{Performance of SERL and SymmGrid.}
\label{tab:eval_comparison}
\small
\setlength{\tabcolsep}{5pt}
\renewcommand{\arraystretch}{1.05}
\begin{tabular}{lcc}
    \toprule
    \textbf{Task} & \textbf{SERL} & \textbf{SymmGrid} \\
    \midrule
    Peg insertion
    & 90.0\% @ 20.13 min
    & 90.0\% @ 15.9 min \\
    Cable routing
    & 93.2\% @ 18.5 min
    & 93.2\% @ 17.0 min \\
    Object relocation
    & 74.7\% @ 10k steps
    & 99.3\% @ 10k steps \\
    \bottomrule
\end{tabular}
\end{table}
\section{Discussion}
SymmGrid achieved significant gains in wall-clock time efficiency compared to SOTA. 
SymmGrid achieved wall-clock training convergence speedups of 1.41x, 2.17x, 1.37x for peg-insertion, cable-routing, and object relocation. It also achieved evaluation success rate improvements of 1.27x, 1.09x for the former two experiments and an 24.6\% performance accuracy gain for object relocation. Note that for cable-routing, SERL did not reach our top performance level, so an estimate cannot be provided. Optimal policy learning occurred as soon as 16.6, 10.9, and 79.3 minutes and on average in 20, 18, and 90 minutes for near 100\% rates. This is remarkable, as on-robot optimal policy learning is close to breaking the 10-minute convergence threshold in two tasks and increase the likelihood of deploying in the wild. The largest nominal time savings came in object relocation, where on average it saved 35.4 minutes and improved evaluation rates by 33.0\% for the forward model. These results confirm that the branching symmetries super-scaling effect for data augmentation plays a significant role in accelerating learning.

In terms of weaknesses, object relocation exhibited a different scaling trend from peg insertion and cable routing: the best performance used three branches and a 0.175 m symmetry workspace. The difference is the use of an exocentric camera. As already mentioned, egocentric scenes are invariant under translational symmetries, while exocentric do not. In scenes that contain multiple depths (table, trays, object, and robot), the use of homographies introduces parallax, interpolation, and border artifacts. Larger symmetry grids increase redundancy among neighboring images and accelerate replay-buffer turnover by inserting many highly correlated samples per physical transition. The larger grids substantially reduce the proportion of unwarped observations in the replay buffer and can shift the buffer's distribution away from real observations, leading the visual encoder to exploit visual artifacts rather than task-relevant geometry. As such, increasing the branch number and the workspace width can eventually reduce the validity and diversity of the augmented data and hurt performance.
\section{Conclusion}
SymmGrid introduced a new paradigm in leveraging symmetry principles for accelerating RL. When DRL trained with branched symmetries via a simple global transformations, the system was able to exploit the globally learned dynamics to perform significantly better at the local level. This work lowers barriers-to-entry for on-robot learning for manipulation and humanoids. Such speedups let us train directly on robots, which is crucial to faithfully learn real contact physics. Such physics can be very unique as shown in this work and would be challenging to learn in simulation. 
\section{Acknowledgments}
This work was supported by the
2024-349547 Chan Zuckerberg Initiative DAF \& managed by the Open Research Community Accelerator (ORCA) \& the Draughon Foundation.
\bibliographystyle{IEEEtran}
\bibliography{references}
\end{document}